\documentclass[conference]{IEEEtran}
\IEEEoverridecommandlockouts
\usepackage{cite}
\usepackage{amsmath,amssymb,amsfonts}
\usepackage{algorithmic}
\usepackage{graphicx}
\usepackage{textcomp}
\usepackage{subfigure}
\usepackage{xcolor}
\usepackage{color, soul}
\usepackage{comment}
\usepackage{svg}
\usepackage{multirow}
\usepackage{tabularx}
\usepackage{dirtytalk}
\def\BibTeX{{\rm B\kern-.05em{\sc i\kern-.025em b}\kern-.08em
    T\kern-.1667em\lower.7ex\hbox{E}\kern-.125emX}}
\begin{document}

\title{Multi-label Video Classification for Underwater Ship Inspection\\

}

\author{\IEEEauthorblockN{Md Abulkalam Azad\IEEEauthorrefmark{1}\IEEEauthorrefmark{2}\IEEEauthorrefmark{3},
Ahmed Mohammed\IEEEauthorrefmark{1},
Maryna Waszak\IEEEauthorrefmark{1}, 
Brian Elvesæter\IEEEauthorrefmark{1}, and
Martin Ludvigsen\IEEEauthorrefmark{3}}
\IEEEauthorblockA{\IEEEauthorrefmark{1}SINTEF AS, Forskningsveien 1, 0373 Oslo, Norway}
\IEEEauthorblockA{\IEEEauthorrefmark{2}Faculty of Sciences and Technology, University of Toulon (UTLN), Toulon, France}
\IEEEauthorblockA{\IEEEauthorrefmark{3}Department of Marine Technology, Norwegian University of Science and Technology (NTNU), Trondheim, Norway}
}

\maketitle

\begin{abstract}
Today ship hull inspection including the examination of the external coating, detection of defects, and other types of external degradation such as corrosion and marine growth is conducted underwater by means of Remotely Operated Vehicles (ROVs). The inspection process consists of a manual video analysis which is a time-consuming and labor-intensive process. To address this, we propose an automatic video analysis system using deep learning and computer vision to improve upon existing methods that only consider spatial information on individual frames in underwater ship hull video inspection. By exploring the benefits of adding temporal information and analyzing frame-based classifiers, we propose a multi-label video classification model that exploits the self-attention mechanism of transformers to capture spatiotemporal attention in consecutive video frames. Our proposed method has demonstrated promising results and can serve as a benchmark for future research and development in underwater video inspection applications.
\end{abstract}

\begin{IEEEkeywords}
Video Classification, Vision Transformer, Underwater Inspection, Deep Learning, Computer Vision
\end{IEEEkeywords}

\section{Introduction}
\subsection{Underwater ship hull inspection}
\begin{figure}[htbp]
    \centering    \includegraphics[width=0.5\textwidth]{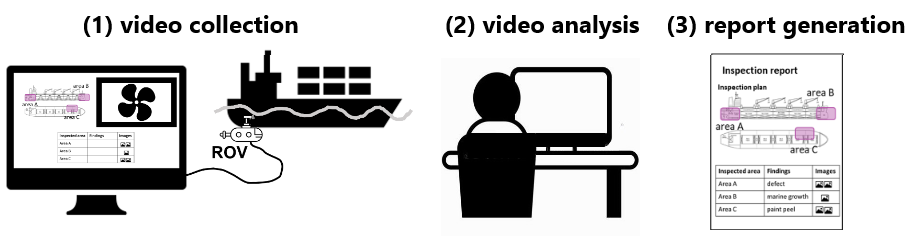}
    \caption{The workflow of current underwater ship inspection using ROVs.}
    \label{fig:process}
\end{figure}

Inspection of marine vessels in the maritime industry plays a significant role in monitoring the life cycle and analyzing the condition of the hull. It examines the external coating and detects potential defects. Corrosion, marine growth, or other external degradation can damage the hull and reduce its lifespan. Ship hull inspections are nowadays shifting to underwater operation from dry-dock to reduce the cost and downtime of the ship. These are conducted by a Remotely Operated Vehicle (ROV) to further cut down the cost and prevent the risk of a human diver. The general procedure as illustrated in Fig. \ref{fig:process} consists of a) collection of videos of the ship hull using an ROV, b) intensive analysis of the videos, and c) preparation of the inspection report. The manual video analysis within the process is time-consuming, tedious, and prone to human error. Therefore, with the advancement of deep learning in computer vision and autonomy in underwater vehicles, automatic video analysis can greatly improve underwater inspection.

\subsection{Frame-wise classification}
A trivial approach to video analysis is to classify each frame of the entire video separately and identify potential threats such as defects or corrosion. This approach only needs an efficient and robust multi-label image classifier and many such off-the-shelf models are available online. We can use a pre-trained image classification model and apply an effective deep transfer learning technique as suggested in \cite{plested2022deep} to fine-tune the model for our domain. A preceding work under the LIACi\footnote{Lifecycle Inspection, Analysis, and Condition information system (https://www.sintef.no/en/projects/2021/liaci/)} project \cite{hirsch2022fusion} also utilized transfer learning to train a multi-label image classifier using the Microsoft Custom Vision \cite{salvaris2018cognitive} framework on the LIACi dataset to classify individual frames in the video. The trained model can predict nine different class labels as illustrated in the methods \& materials section on the surface of the ship hull. However, this approach has a significant limitation as it only considers spatial information from static image frames and lacks the temporal insight that is essential for Video Understanding \cite{huang2018makes}. As a result, the model becomes temporally unstable.

\subsection{Main objective}
In order to alleviate the issue, it is necessary to train a model by learning spatiotemporal information from videos which can improve the automatic video analysis of underwater ship hull inspections. Unlike temporal action recognition and localization \cite{xia2020survey} that consider dynamic foreground and background objects, our  videos only have static scenes including ROV motion with a dynamic camera. Hence, the benefit of utilizing the temporal aspects can facilitate stabilization during the video analysis. Our core focus is to enhance the consistency and stability of the model's predictions during underwater video analysis. Therefore, in this paper, we investigate the consistency and stability of image-based classifiers which can help us in understanding the advantages and limitations of using an image-based multi-label classifier for this purpose. Furthermore, we propose a video classification model that takes into account both temporal and spatial information. In summary, the contributions of this work are;
\begin{enumerate}
    \item[a.] Analysis of image-based classifiers (benefits and limitations).
    \item[b.] Exploration of the benefits of adding temporal information.
    \item[c.] Identification of a deep learning multi-label video classifier for labeling video frames based on spatiotemporal attention.
\end{enumerate}

\vspace{0.5cm}
The rest of the paper is divided into five sections. Related works are described in section \ref{related_work}, whereas section \ref{MnM} unveils the methods \& materials we utilize within this work. Sections \ref{result} and \ref{analysis} include the results of our works and ablation study. Finally, we conclude in section \ref{conclusion} by leaving some discussion and direction for further research and development in the same area. 

\section{Related works}\label{related_work}
\subsection{Computer vision technology}
Computer vision has been used in automating various industries worldwide. While artificial intelligence enables machines to think, computer vision provides them with the ability to see. It has been used in many diverse fields such as agriculture, autonomous vehicle, facial recognition, medical imaging, manufacturing, and many more. Convolutional Neural Network (CNN) is widely recognized as a breakthrough innovation in this area which was introduced in 1998 \cite{lecun1998gradient} for hand-written digit recognition tasks from images. CNN extracts spatial information from images which helps with the recognition and classification tasks. Since then, several groundbreaking innovations \cite{simonyan2014very, krizhevsky2017imagenet, szegedy2015going} have been achieved to improve this technology further. Therefore, utilizing a CNN-based architecture to extract spatial features from video frames is a valuable addition to automatic underwater video analysis.

\subsection{Vision Transformer (ViT)}
Following the immense success of the Self-attention based Transformer \cite{vaswani2017attention} in the field of Natural Language Processing (NLP), it has also evolved in a wide range of applications within Computer Vision. Researchers thrived to adapt the self-attention mechanism in the Computer Vision area and introduced the Vision Transformer \cite{dosovitskiy2020image} in 2020 as the counterpart of the original Transformer. ViT addresses image recognition tasks by dividing an input image into patches and applying self-attention to these patches to obtain spatial contextual relations between them. Thus, it has been adapted together with traditional CNN architectures for image recognition tasks \cite{graham2021levit,liu2021swin,wang2021pyramid}. The revolution of the ViT has also shifted through different variations to other vision tasks including object detection \cite{carion2020end, dai2021up}, and image segmentation \cite{li2022panoptic}.

We are particularly interested to train a multi-label ViT image classifier on LIACI dataset because of its outstanding self-attention mechanism. This facilitates better spatial feature extraction on frames during video analysis. ViT applies a standard NLP-suited transformer on an image which is first split into fixed-size patches in order to make the fewest possible adjustments. The list of patches is similar to tokens or words of NLP applications which are fed to the transformer network as inputs. This approach is called patch embedding. In order to get positional information, standard 1D position encoding is added along with the input sequence of patches. The rest of the architecture is designed by the transformer encoder layers where a learnable embedding is prepended to the embedded patches sequence. One major limitation of ViT is that it needs to be pre-trained on large-scale datasets and then fine-tuned on smaller datasets to surpass CNN for downstream tasks. While pre-training, a Multi-layer perceptron (MLP) based classification head is integrated with one hidden layer. The MLP layer is later replaced by one single linear layer during fine-tuning. Recently, a study \cite{chen2021vision} has shown that ViT can outperform CNN models of similar size when trained on ImageNet from scratch without strong data augmentations which overcome the large-scale pretraining limitation. Therefore, it is apparent that ViT holds promises for the underwater video analysis domain as well.

\subsection{Temporal Action Localization (TAL)}
To study video understanding, we need to start with extracting temporal information from the frames of a video. Temporal Action Localization (TAL) \cite{xia2020survey} refers to determining the time intervals in a video that contains a target action. The target action is usually a dynamic activity (e.g., marine plant waving, fish swimming) but can be a stationary fact as in our case which lasts for an indefinite duration such as corrosion in a ship hull. TAL mainly performs two tasks; recognition and localization. Recognition denotes the detection of the class labels whereas localization determines the start and end time of the detected actions. The latter does not apply to our work at the moment as we only focus on multi-label class recognition.

Generally, there are two types of TAL methods: single-stage and two-stage; single-stage: generates several temporal action segments (start to end) proposals in an untrimmed long video and classifies these actions simultaneously, two-stage: first proposes segments and classifies actions and then regresses the boundaries. In addition, there are a couple more variations depending on the data annotations; 
\begin{itemize}
    \item \textbf{Fully-Supervised Temporal Action Localization (F-TAL):} It refers to the training when the dataset contains both the video-level category classes and the temporal annotations (start and end time) of the action segments.
    \item \textbf{Weakly-Supervised Temporal Action Localization (W-TAL):} In the realistic scenario, most of the videos are untrimmed with no temporal information and contain many frames that are not relevant to target actions. So it is very difficult to acquire temporal annotations.
\end{itemize}
W-TAL indeed coincides with our case as we have only untrimmed underwater videos without annotations. However, the implementation of video classification requires video annotation. This needs extensive time to prepare the data for training a deep learning video classifier. Hence, we follow a similar W-TAL approach to train our multi-label video classifier.

\subsection{Spatiotemporal features in video classification}
In video understanding, the improved Dense Trajectories (iDT) proposed in \cite{wang2013action} was the state-of-the-art hand-crafted feature for classification tasks. The iDT descriptor demonstrates the ability to extract temporal features differently from that spatial information. Consequently, 3D ConvNets was proposed in \cite{tran2015learning} to learn spatiotemporal features from videos. It also overcomes the limitation of 2D ConvNets which loses temporal information of the input signal right after every convolution operation. The best architecture proposed in their experiment, called C3D net, is homogeneous and comprises 8 convolution, 5 max-pooling, and 2 fully connected layers, followed by a softmax output layer. The 3D convolution kernels in this network are 3x3x3 with a stride of 1 in both spatial and temporal dimensions. They also claimed that a trained C3D network can serve as a potential spatiotemporal feature extractor for other video analysis tasks which could be advantageous in our scenario.

TimeSformer \cite{bertasius2021space} is among the first video models to incorporate self-attention mechanisms in video understanding inspired by the success of self-attention mechanisms in ViT. It utilizes self-attention over both spatial and temporal dimensions of an input video sequence rather than using 3D CNN to extract temporal features along the frames. The model takes an input snippet consisting of 8 RGB frames of size 224x224, decomposes each frame into 16x16 patches, and applies self-attention along the temporal patches for these 8 consecutive frames. During inference, it uses 3 spatial crops from the temporal clip and predicts by averaging the scores. In contrast to our approach of using consecutive frames to predict static class labels in the current frame, TimeSformer samples the 8 frames of an input video at a rate of 1/32, and these frames are not necessarily consecutive. Their experiments have demonstrated that the best performance is achieved when temporal and spatial attention are applied separately. Adopting this approach will be crucial in training our model video classifier. 

ViViT \cite{arnab2021vivit} is another example of a transformer-based video classification model that benefits from the self-attention mechanism. They propose four variations of their model by factorizing the spatial and temporal dimensions in different ways, ranging from simple to complex architectures. They also explain how to utilize pre-trained ViT image models to train a video classifier on small datasets along with effective regularization techniques which could be particularly advantageous for our purposes. They emphasize the operational flexibility of a variable number of input frames which is similar to the original transformer's ability to handle any sequence of input tokens. While there are similarities with TimeSformer \cite{bertasius2021space}, the rich ablation study presented in ViViT provides a strong foundation for us to begin with our own video model.

In essence, the video models based on 3D CNN or transformers provide a promising research direction for developing a suitable multi-label video classifier for underwater ship inspection. Although the underlying architecture of our model will be similar to these models, it will serve a different purpose. Our model will predict static classes instead of dynamic actions by absorbing the disrupted motions in the video and will stabilize the prediction confidence along the temporal dimension.

\section{Materials \& methods}\label{MnM}
\subsection{Datasets}
\begin{figure}[htbp]
    \centering    \includegraphics[width=0.5\textwidth]{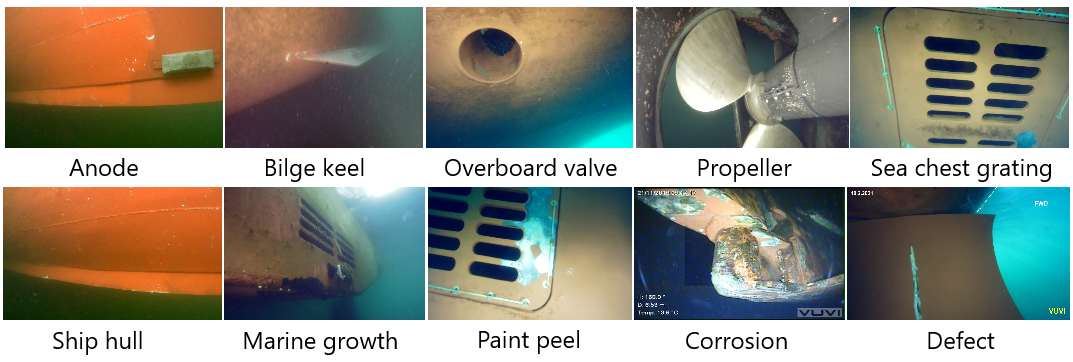}
    \caption{Visualization of 10 class labels of two different categories.}
    \label{fig:class_labels}
\end{figure}

The LIACI dataset for underwater ship Lifecycle Inspection, Analysis, and Condition Information is publicly available and has been published in \cite{waszak2022semantic}. The dataset comprises 1893 RGB images extracted from 17 inspection videos of various ships. There are 10 class labels as depicted in Fig. \ref{fig:class_labels} divided into two different categories;
\begin{itemize}
    \item \textbf{Ship components:} \textit{Anode}, \textit{Bilge keel}, \textit{Overboard valve}, \textit{Propeller}, \textit{Sea chest grating}, and \textit{Ship hull}.
    \item \textbf{Common marine coating issues:} \textit{Marine growth}, \textit{Paint peel}, \textit{Corrosion}, and \textit{Defect}.
\end{itemize}

\begin{figure}[htbp]
    \centering    \includegraphics[width=0.48\textwidth]{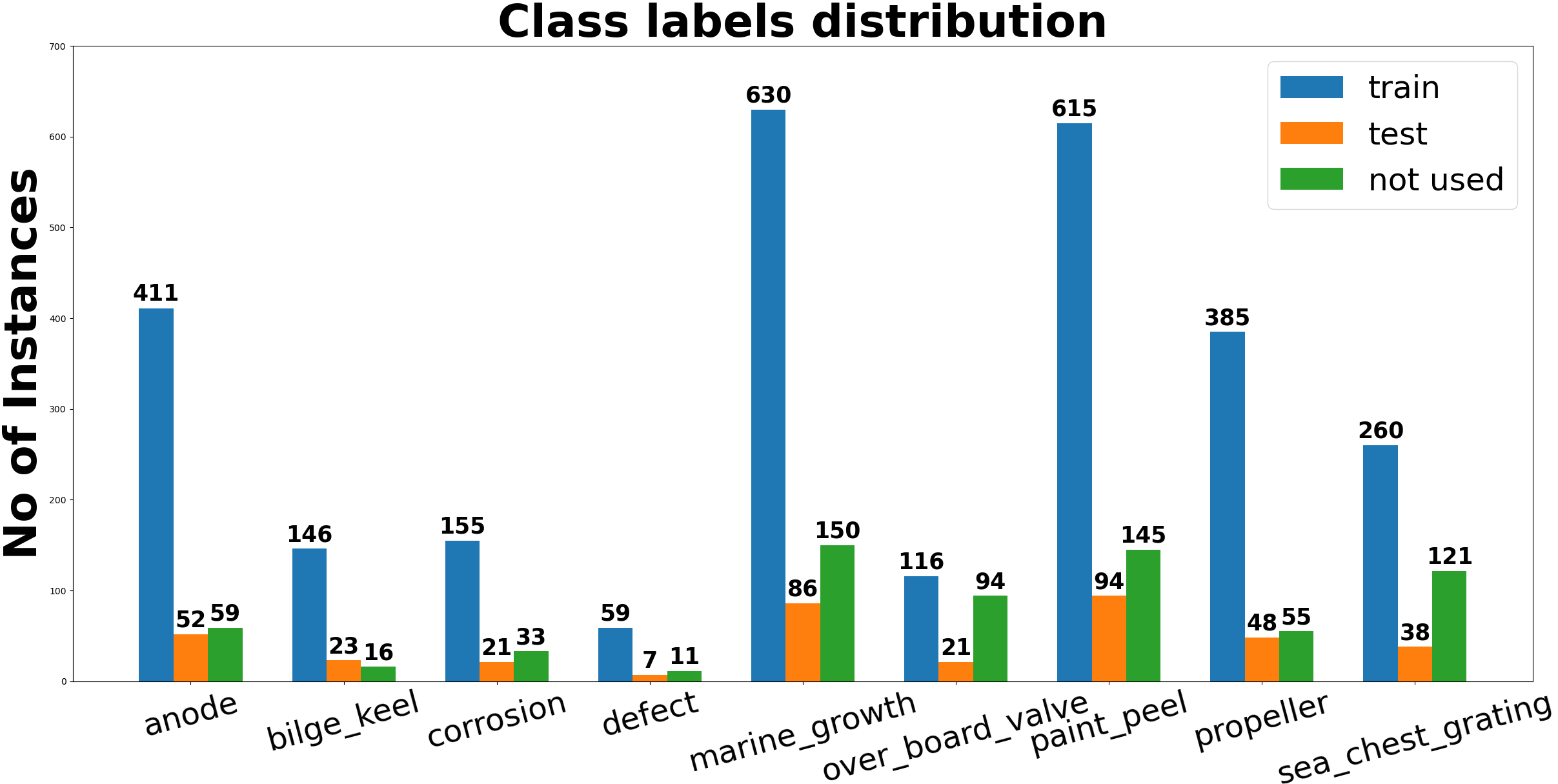}
    \caption{Distribution of class instances.}
    \label{fig:data_dist}
\end{figure}

However, we exclude the \textit{Ship hull} class during the training of our deep learning model as it is present in all images. We only used 1561 images from the LIACI dataset to train and test our model as recommended by the authors \cite{waszak2022semantic}. The remaining 332 images were considered too spatially similar to other images in the dataset (Cosine similarity cut-off of 0.90). The class instance distribution in Fig. \ref{fig:data_dist} indicates that while the dataset is not perfectly balanced, it is not severely imbalanced either.

Furthermore, to comprehensively analyze and evaluate the performance of trained models, we selected 8 key clips of 1920x1080 resolution from an underwater inspection video.
These clips were chosen randomly from untrimmed inspection videos and each clip is approximately 14 seconds long. Table \ref{tab:video_clips} provides descriptions of the physical content of the clips that are easily recognizable to human eyes. However, distinguishing between \textit{marine\_growth}, \textit{corrosion}, and \textit{paint\_peel} with human visual perception can be quite challenging most of the time. The results of the analysis and evaluation are documented in sections \ref{result} and \ref{analysis}.
\begin{table}[htpb!]
    \centering
    \caption{Contents of the 8 key video clips.}
    \begin{tabular}{c | c}
        \hline
        Serial & Major physical real contents  \\
        \hline
        1 & anode, paint\_peel\\
        2 & bilge\_keel, paint\_peel, over\_board\_valve, anode\\
        3 & propeller, paint\_peel, corrosion, marine\_growth\\
        4 & paint\_peel, marine\_growth, propeller\\
        5 & marine\_growth, propeller, corrosion\\
        6 & paint\_peel\\
        7 & propeller, marine\_growth\\
        8 & sea\_chest\_grating, paint\_peel, corrosion\\
    \end{tabular}
    \label{tab:video_clips}
\end{table}

\subsection{Multi-label ViT Image Classifiers}
In \cite{dosovitskiy2020image}, a few variants of ViT models are proposed that differ in model size and input patch size. For instance, the ViT-L/16 refers to the \say{Large} variant and is composed of 24 training layers with a 16x16 input patch size. The PyTorch \cite{paszke2019pytorch} vision package includes several ViT models that can be easily implemented. Besides, PyTorch enables access to the models' underlying architecture and allows us to modify them through retraining or fine-tuning conveniently. Based on the model's capacity, our requirements, and computing resources we selected the ViT-B/16 architecture. The size of the model is 330.3MB with 86M trainable parameters and it has 95.318\%@5 accuracy on ImageNet 1K dataset \cite{deng2009imagenet}.

We decided to train two versions of the ViTs on LIACI data with pre-trained on ImageNet 1k and COCO 2014 \cite{lin2014microsoft} datasets respectively and compare their performances. Although the ImageNet pre-trained ViT is readily available in PyTorch, we need to train the COCO version by ourselves in advance. We downloaded the COCO dataset using FiftyOne \cite{moore2020fiftyone} and fully finetuned an ImageNet pre-trained ViT model on COCO. Finally, we trained our two desired ViT models pre-trained from ImageNet and COCO datasets and abbreviated them as IMAGENET\_ViT and COCO\_ViT respectively. The training hyperparameters are the same for both as shown in Table \ref{tab:init_param} along with the data transformations. It is noted that we applied separate image normalization by computing respective mean (M) and standard deviation (S) on LIACI and COCO datasets. Also, only the Image Resize and Normalization are applied during validation or evaluation. Nonetheless, we investigated various hyperparameters and data augmentations that are exhibited in section \ref{analysis}.
\begin{table}[htbp]
\caption{Training hyperparameters and data transformations for IMAGENET\_ViT and COCO\_ViT}
\begin{center}
\begin{tabularx}{\linewidth}{|X|c|c|}
\hline
\textbf{Type} & \textbf{\textit{IMAGENET\_ViT}}& \textbf{\textit{COCO\_ViT}}\\
\hline
\textbf{Loss function} & BCEWithLogitsLoss & BCEWithLogitsLoss\\
\hline
\textbf{Optimizer} & SGD & SGD\\
\hline
\textbf{Learning rate} & 0.001 & 0.001\\
\hline
\textbf{Momentum} & 0.9 & 0.9\\
\hline
\textbf{Batch size} & 16 & 16\\
\hline
\multirow{2}{*}{\textbf{Scheduler}} & StepLR & StepLR\\
 & (step=20, gamma=0.1) & (step=20, gamma=0.1)\\
 \hline
\multicolumn{3}{|c|}{\textbf{Data Transformations}}\\
\hline
\textbf{Image Resize} & 224x224 & 224x224\\
\hline
\textbf{Normalization} & M[0.348, 0.369, 0.352] & M[0.348, 0.369, 0.352]\\
\textbf{(LIACI Data)} & S[0.249, 0.244, 0.206] & S[0.249, 0.244, 0.206]\\
\hline
\textbf{Normalization} & N/A & M[0.485, 0.456, 0.406]\\
\textbf{(COCO Data)} & N/A & S[0.229, 0.224, 0.225]\\
\hline
\textbf{Random} & \multirow{2}{*}{p=0.5} & \multirow{2}{*}{p=0.5}\\
\textbf{Horizontal Flip} & &\\
\hline
\end{tabularx}
\label{tab:init_param}
\end{center}
\end{table}

\subsection{Prediction Confidence and Temporal Characteristics}
\begin{figure}[htbp]
    \centering    \includegraphics[width=0.48\textwidth]{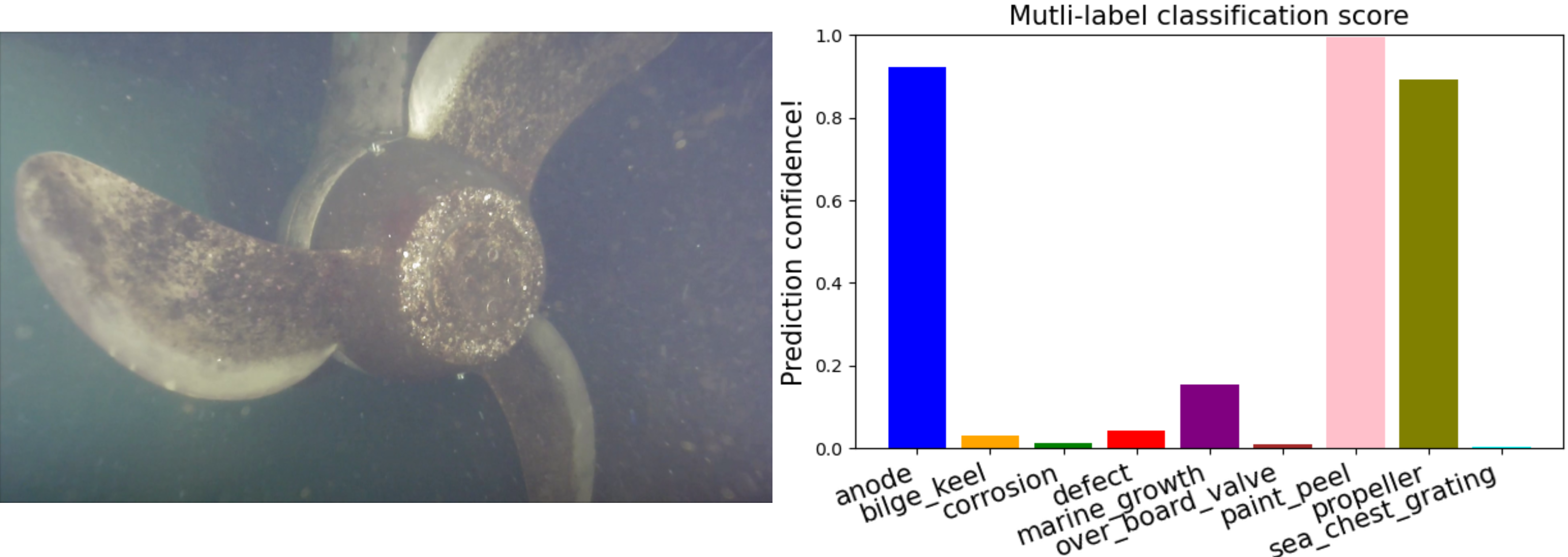}
    \caption{Model's prediction confidence on a frame during a video inspection.}
    \label{fig:frame_prediction}
\end{figure}

To analyze a trained model's confidence behavior, we leverage OpenCV \cite{opencv_library} to process a video snippet and observe the model's prediction confidence on each frame, as illustrated in Fig. \ref{fig:frame_prediction}.
This approach also enabled us to evaluate a model's ability to predict multiple class labels simultaneously on a per-frame basis.

To integrate temporal reasoning into our model, it is necessary to examine and analyze the model's temporal consistency throughout the development process. To achieve this, we utilize OpenCV to observe the temporal aspect of the model's confidence for different labels during an inspection. This is useful to qualitatively assess the temporal stability of a trained model and is depicted in the result section.  


\subsection{Underwater Image Quality Metrics}
In underwater image or video tasks, measuring image quality is a grave concern as it directly impacts any vision-based operation. Poor-quality images can significantly degrade the performance. To measure frame quality, we employed two separate image quality metrics - UCIQE \cite{yang2015underwater} and UIQM \cite{panetta2015human} - to establish a correlation between the model's prediction confidence and frame quality. Both metrics are no-reference and meticulously designed for underwater images. The qualitative output of these two metrics is reported in the result section. 

\subsection{Video data Generation and Annotation}
We have acquired the corresponding videos of LIACI training images which are untrimmed and unstructured video data. We were able to extract 755 corresponding video snippets out of 1893 images contained in the dataset. Each snippet consisted of seven consecutive frames, with the middle frame representing the original image from the LIACI dataset and its class labels considered as the labels for the entire snippet during training. This approach may be considered a weakly supervised data annotation. The snippets were split into 584 for training, 87 for validation, and 84 that were not used by following the same splitting convention of the image dataset. It is worth noting that the generated video dataset contains fewer snippets than half of the number of images in the LIACI dataset. As a result, it may not be sufficient to train a robust video model compared to the image model. 

\subsection{Multi-label Video Classifiers}
\begin{figure}[htbp]
    \centering    \includegraphics[width=0.5\textwidth]{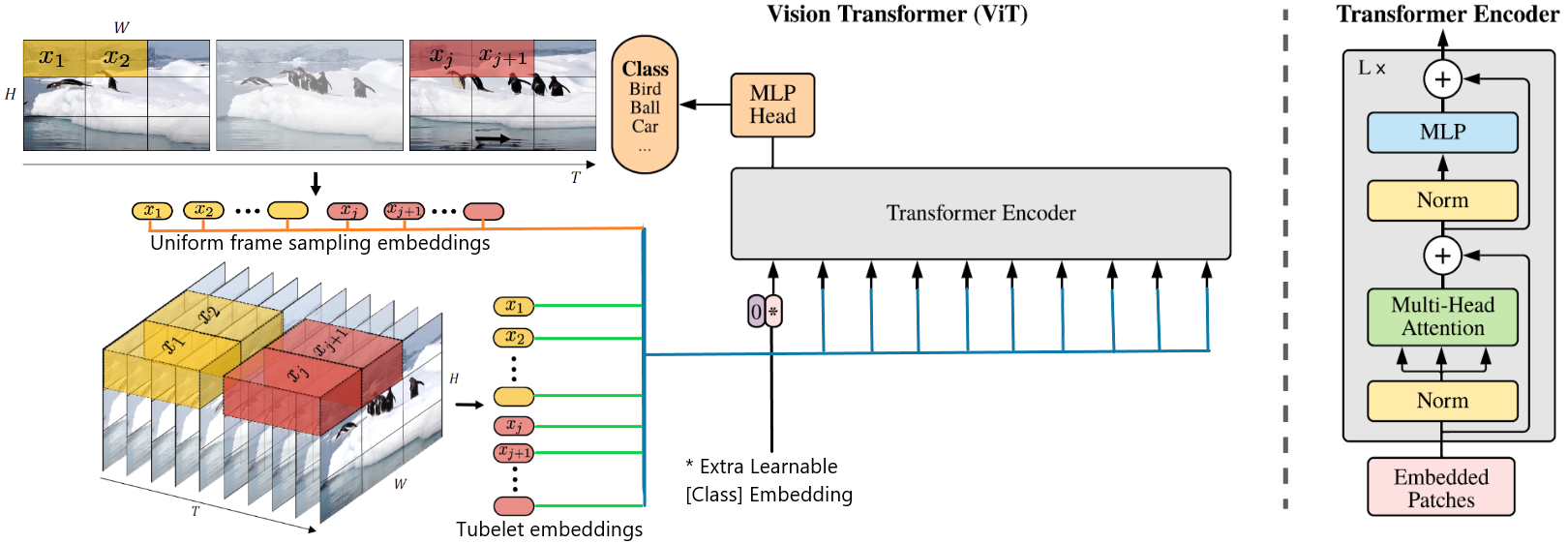}
    \caption{A simple approach to video model using the same architecture as the image classifier.}
    \label{fig:video_model_1}
\end{figure}

We have implemented and trained 6 different variants of ViT-based multi-label video classifiers. Initially, we adopted a straightforward method by utilizing the spatiotemporal token embedding techniques proposed in \cite{arnab2021vivit}. We trained our first 2 variants by extracting tokens from the video snippets using either uniform frame sampling or tubelet embedding methods, and then input these tokens directly into a base ViT encoder. The process is illustrated in Fig. \ref{fig:video_model_1}, and the diagrams used are borrowed from \cite{arnab2021vivit} and \cite{dosovitskiy2020image}. To implement uniform frame sampling, we extracted 28 patches with dimensions of 32x56 from each frame of a seven-frame input snippet, generating a total of 196 patch embeddings. These embeddings are readily compatible with a base ViT architecture. On the other hand, to achieve the tubelet embedding as depicted in Fig.\ref{fig:video_model_1}, we utilized a pretrained 3D ResNet18 model to extract C3D features from the input snippet.

The rest of the 4 video classifiers are implemented by applying different underlying strategies based on Model 2 proposed in \cite{arnab2021vivit} which is similar to the TimeSformer method presented in \cite{bertasius2021space}. This approach uses a ViT base architecture called a spatial transformer encoder to extract spatial features from each frame. These consecutive spatial features are then passed through a temporal transformer to combine with temporal features, followed by an MLP head to predict class labels. This method is designed to address the issue of overfitting on smaller datasets such as ours and provides a more sophisticated model for video classification. A previously trained ViT image classifier is adopted as the spatial transformer encoder, while a new standard transformer is employed as the temporal one. During training, we froze the weights of the spatial transformer and solely updated the temporal transformer. This approach resulted in a notable acceleration of the training process and facilitated the adaptation of the models to finetuning tasks. 

\subsection{Multi-label Evaluation Metrics}
The computation of multi-label classification evaluation metrics is different from multi-class classification. The Scikit-learn Python package \cite{scikit-learn} provides essential tools to easily compute different metrics. We report accuracy, precision, recall, and f1-score on the validation set of LIACI data for our image and video models in section \ref{result}. These metrics are calculated along the instances and averaged over them. The mathematical equations are as follows in Eq. \eqref{accuracy}, \eqref{precision}, \eqref{recall}, and \eqref{f1-score} where \(n\) is the number of images, \(y\) is the ground truth, and \(\hat{y}\) is the predicted label. Besides, we computed class-wise evaluation metrics during some analysis in section \ref{analysis}.

\begin{equation}
Accuracy = \frac{1}{n} \sum_{i=1}^n \frac{|y_i \cap \hat{y}_i|}{|y_i \cup \hat{y}_i|} \label{accuracy}
\end{equation}
\begin{equation}
Precision = \frac{1}{n} \sum_{i=1}^n \frac{|y_i \cap \hat{y}_i|}{|\hat{y}_i|} \label{precision}
\end{equation}
\begin{equation}
Recall = \frac{1}{n} \sum_{i=1}^n \frac{|y_i \cap \hat{y}_i|}{|y_i|} \label{recall}
\end{equation}
\begin{equation}
F1-score = \frac{1}{n} \sum_{i=1}^n \frac{2|y_i \cap \hat{y}_i|}{|y_i| + |\hat{y}_i|} \label{f1-score}
\end{equation}

\subsection{Hardware Resources}
We used NVIDIA RTX 2080 Ti (11GB) and RTX A6000 (48GB) GPUs to train both of our image and video models. For inference and testing, we used a local system that constitutes of NVIDIA GTX 980 (4GB) with Intel(R) Xeon(R) CPU E5-1650v3 @3.50GHz and 32GB RAM.

\section{Results}\label{result}
\begin{figure}[htbp]
    \centering    \includegraphics[width=0.48\textwidth]{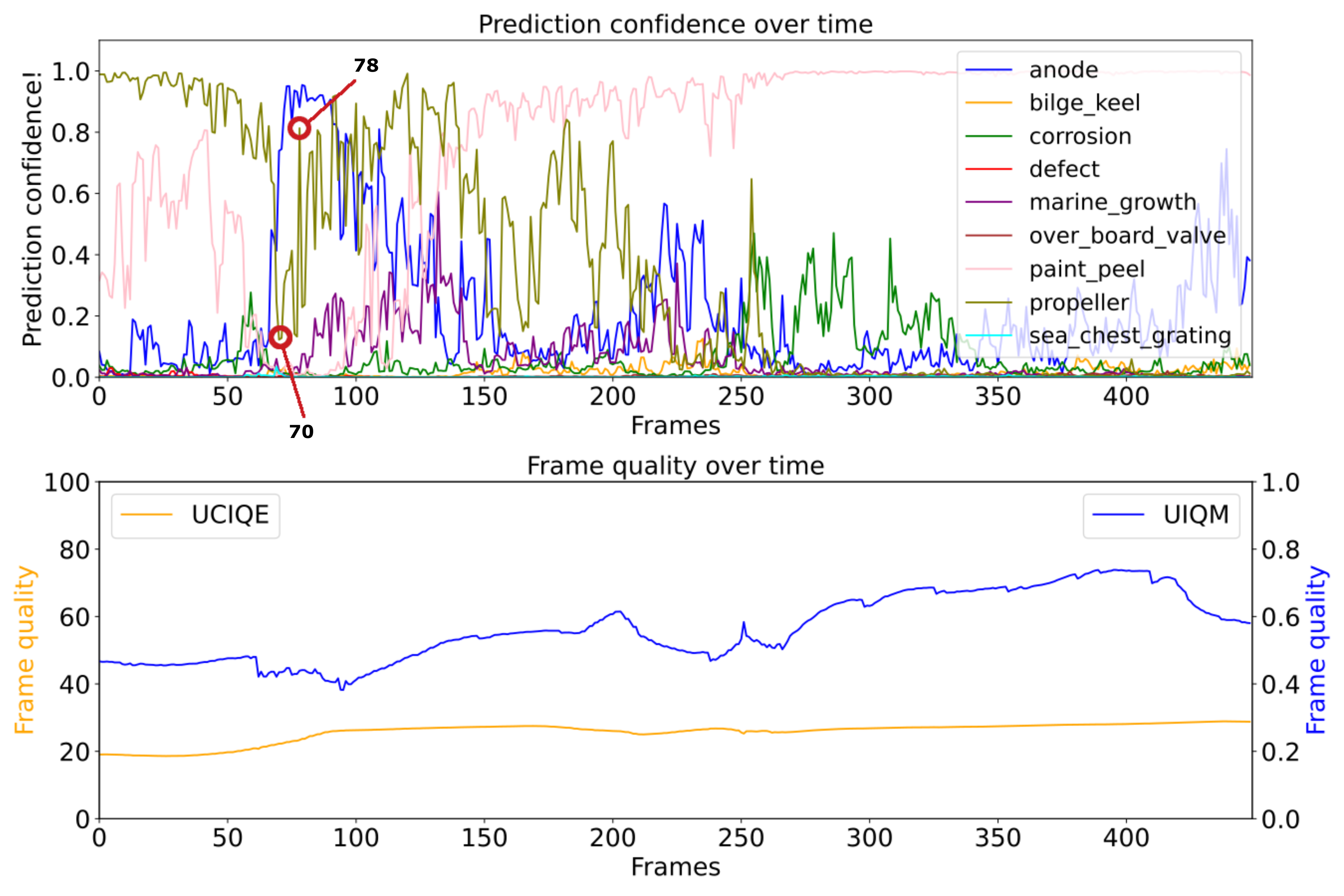}
    \caption{Temporal observation with UCIQE and UIQM metrics on a video snippet.}
    \label{fig:temporal_quality}
\end{figure}

\begin{figure}[htbp]
     \centering
     \begin{subfigure}
         \centering
         \includegraphics[width=0.23\textwidth]{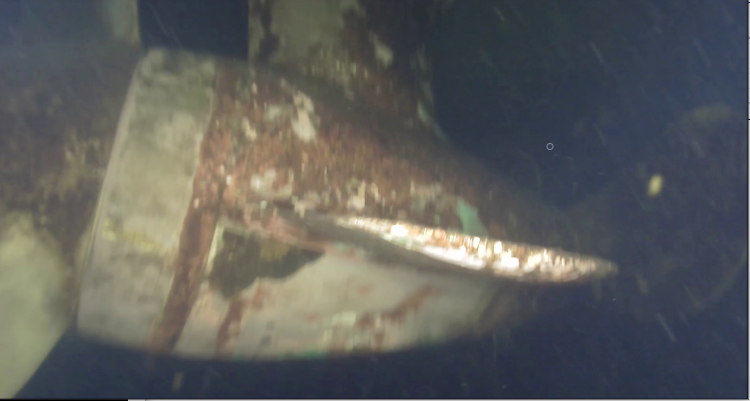}
     \end{subfigure}
     \hfill
     \begin{subfigure}
         \centering
         \includegraphics[width=0.23\textwidth]{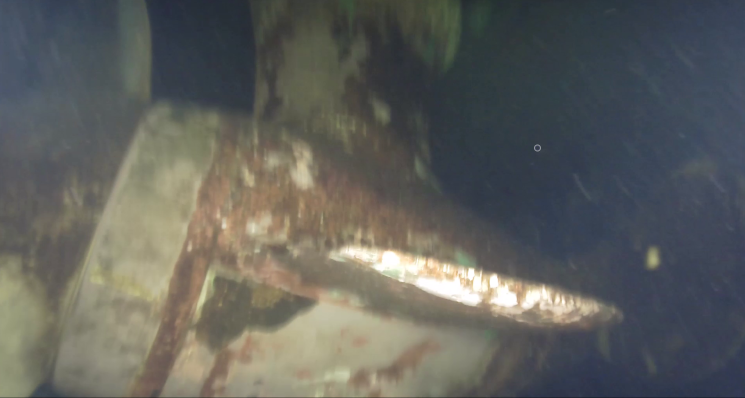}
     \end{subfigure}
    \caption{Frame 70 and 78 (left to right) of a video snippet.}
    \label{fig:frames}
\end{figure}

The temporal observation of video clip no.3 from Table \ref{tab:video_clips} is illustrated in Fig. \ref{fig:temporal_quality} using a model trained on the LIACI dataset through Microsoft Custom Vision in \cite{hirsch2022fusion}. Although the model successfully detects a couple of classes, the confidence values for consecutive frames fluctuate significantly. We noticed similar behavior for other snippets even though the spatial changes between frames are negligible. The bottom row of Fig. \ref{fig:temporal_quality} displays the output of the two image quality metrics mentioned in section \ref{MnM} on a single video clip, while comparing them against a model's temporal prediction confidence.
Since UCIQE and UIQM have different value ranges, we plot these metrics on two different scales within the same plot. Consequently, it is evident that UCIQE does not exhibit any correlation with the observed fluctuation, whereas UIQM indicates that the prediction tends to be consistent with higher UIQM values between frames 250 to 450. On the other hand, the highlighted confidence values for frames 70 and 78, differ significantly at 0.12 and 0.81, respectively, despite a negligible spatial difference between them, as shown in Fig. \ref{fig:frames}. Therefore, the rest of this section demonstrates to what extent our image and video models gradually overcome the issue.

\subsection{IMAGENET\_ViT and COCO\_ViT Image Classifers}
\begin{figure}[htbp]
    \centering    \includegraphics[width=0.48\textwidth]{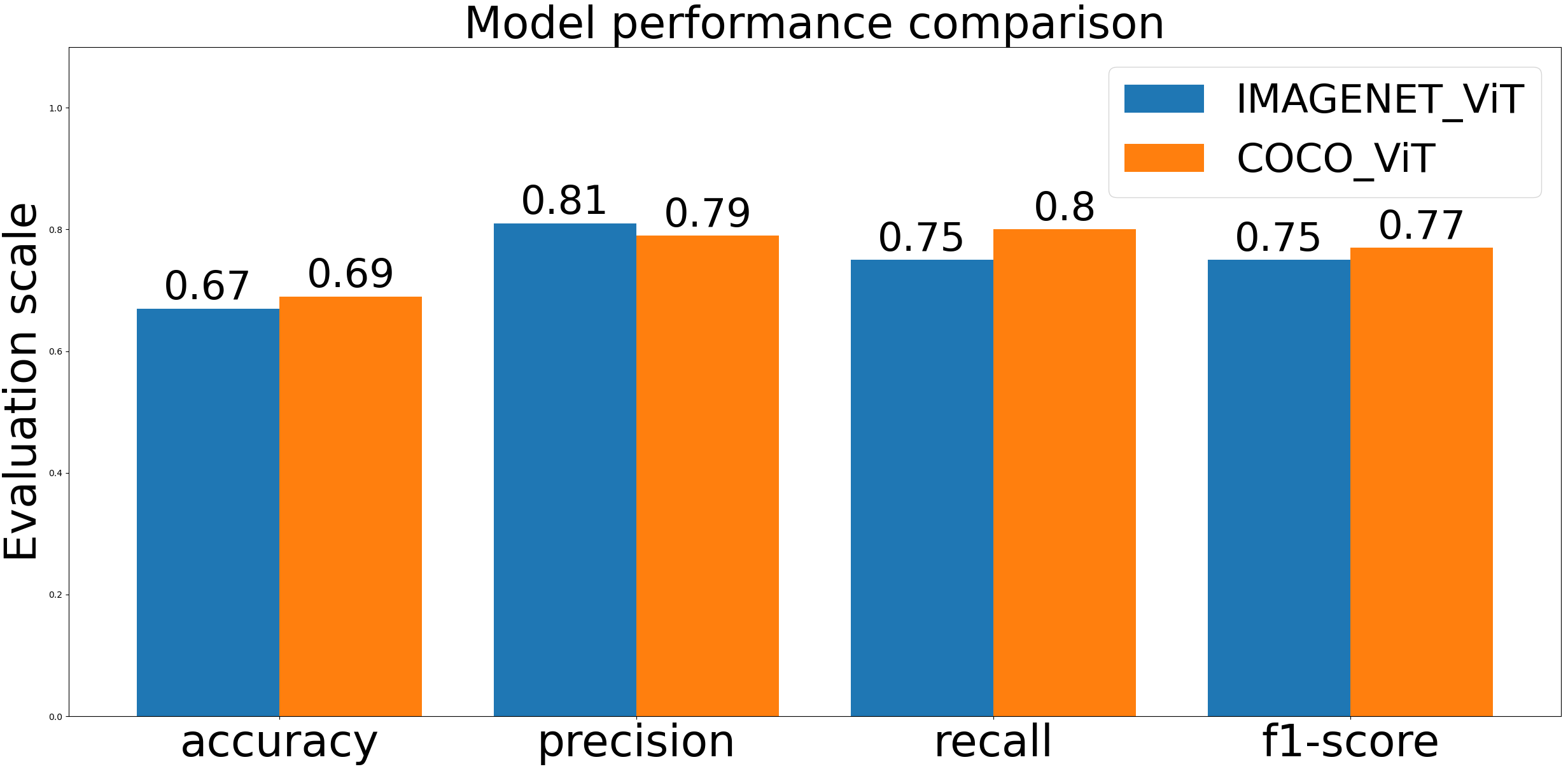}
    \caption{Evaluation metrics comparison between our IMAGENET\_ViT and COCO\_ViT models on the validation dataset.}
    \label{fig:eval_mets}
\end{figure}

Once we began the training process using the hyperparameters and transformations outlined in section \ref{MnM}, we conducted an extensive analysis to determine the optimal models. Consequently, we found the best performances by utilizing the hyperparameters and transformations presented in Table \ref{tab:final_param}. A comparative quantitative evaluation for both of our models is shown in Fig. \ref{fig:eval_mets}.
\begin{table}[htbp]
\caption{Optimal hyperparameters and transformations for IMAGENET\_ViT and COCO\_ViT}
\begin{center}
\begin{tabularx}{\linewidth}{|X|c|}
\hline
\multicolumn{2}{|c|}{\textbf{Hyperparameters}}\\
\hline
\textbf{Loss function} & BCEWithLogitsLoss \\
\hline
\textbf{Optimizer} & SGD \\
\hline
\textbf{Learning rate} & 0.001\\
\hline
\textbf{Momentum} & 0.9 \\
\hline
\textbf{Batch size} & 16 \\
\hline
\multirow{2}{*}{\textbf{Scheduler}} & ReduceLROnPlateau \\
 & mode=’min’, factor=0.1 \\
 \hline
\multicolumn{2}{|c|}{\textbf{Data Transformations}}\\
\hline
\textbf{Image Resize} & 224x224 \\
\hline
\multirow{2}{*}{\textbf{Normalization}} & M[0.348, 0.369, 0.352] \\
& S[0.249, 0.244, 0.206] \\
\hline
\textbf{GaussianBlur} & kernel\_size=(5, 9), sigma=(0.1, 5), p=0.5 \\
\hline
\textbf{AugMix() \cite{hendrycks2019augmix}} & p=0.5 \\
\hline
\textbf{Random} & \multirow{2}{*}{p=0.5} \\
\textbf{Horizontal Flip} &\\
\hline
\end{tabularx}
\label{tab:final_param}
\end{center}
\end{table}
While both models exhibit almost similar performances in each evaluation metric, COCO\_ViT outperformed IMAGENET\_ViT by a small margin in all metrics except precision.

\begin{figure}[htbp]
    \centering    \includegraphics[width=0.48\textwidth]{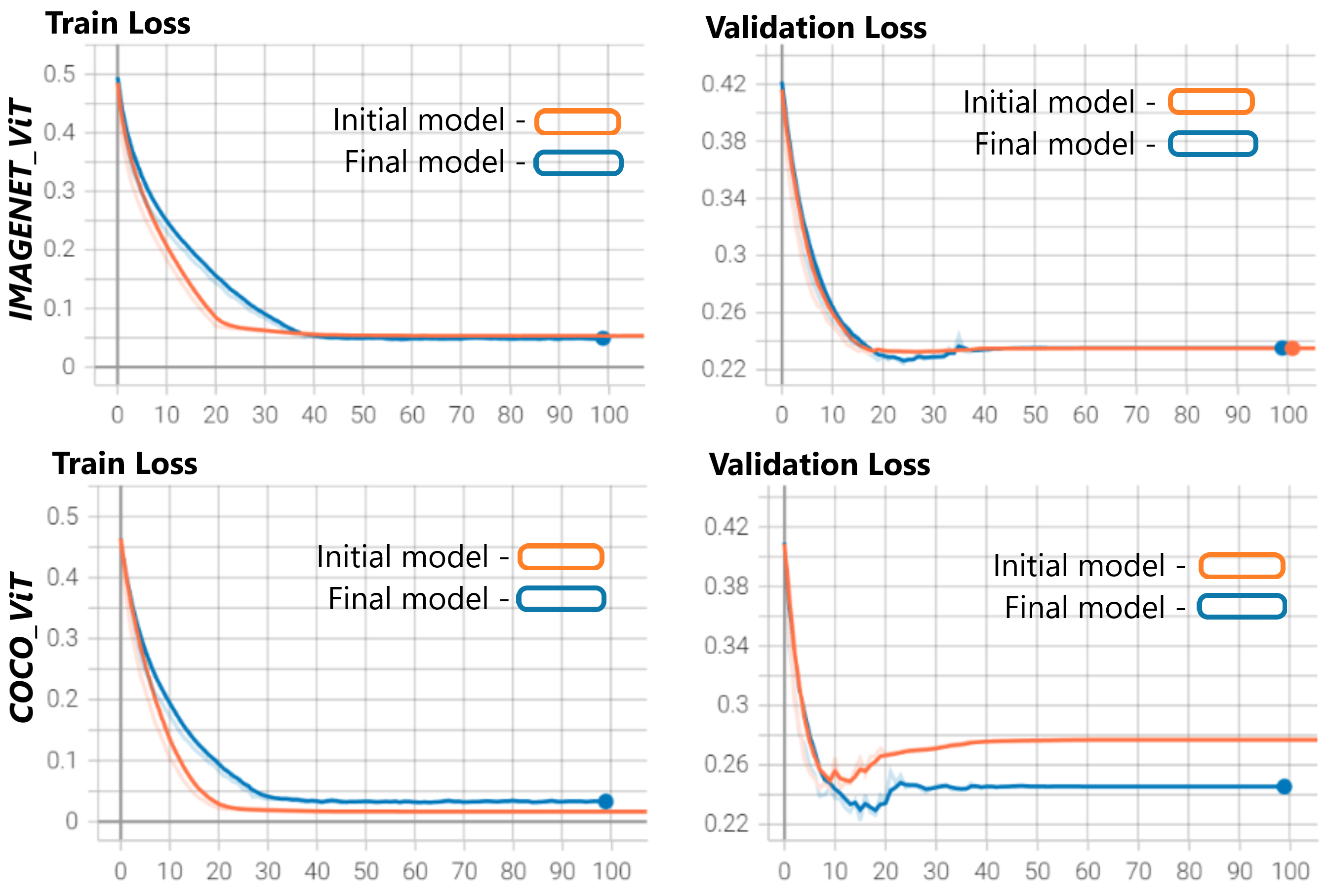}
    \caption{Comparison between the initial and final models in finding local minima for the loss during training. The optimal validation loss for IMAGENET\_ViT is within 20 to 30 epochs. COCO\_ViT exhibits more regularization than the initial model.}
    \label{fig:loss_init_final}
\end{figure}

\begin{figure}[htbp]
    \centering    \includegraphics[width=0.48\textwidth]{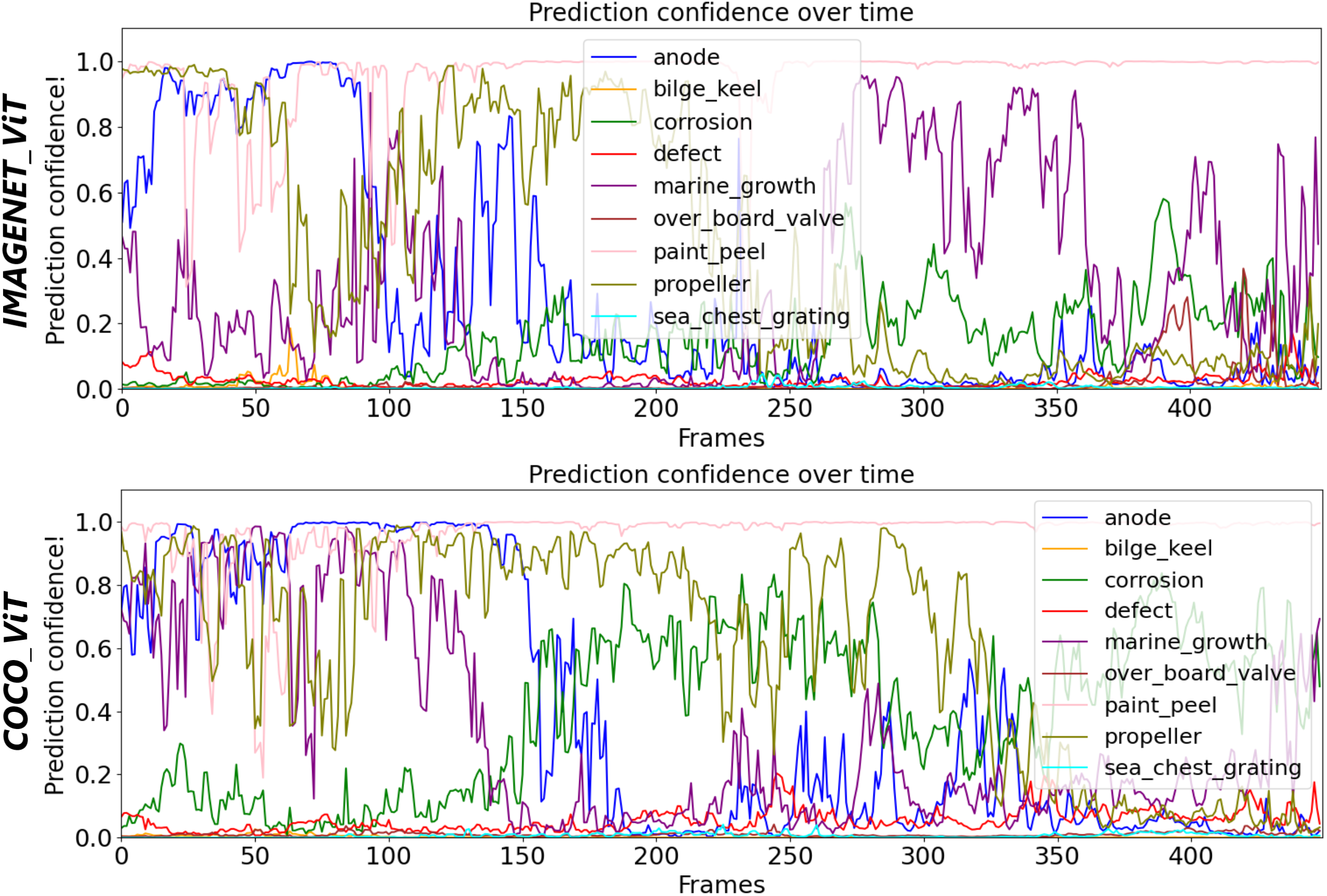}
    \caption{Temporal observation of the final IMAGENET\_ViT and COCO\_ViT on the same video snippet as in Fig. \ref{fig:temporal_quality}.}
    \label{fig:temporal_init_final}
\end{figure}

The ReduceLROnPlateau learning rate scheduler aids in finding better local minima on the validation loss. Fig. \ref{fig:loss_init_final} shows that the final model was able to find the minimal loss on validation compared to the initial one in both cases. Increasing the loss of the final model during training compared to the initial model and subsequently reducing the loss more on the validation set leads to better regularization of COCO\_ViT. On the other hand, the utilization of Gaussian blur and AugMix \cite{hendrycks2019augmix} enhanced the stability of the model's confidence in temporal analysis by facilitating the learning of abrupt ROV motion during inspections. Hence, Fig. \ref{fig:temporal_init_final} demonstrates that both models improved the stability of temporal confidence, particularly in detecting the \textit{Paint peel} class, in contrast to Fig. \ref{fig:temporal_quality}. Furthermore, the models exhibited a more exploratory nature in detecting other class labels during the inspection which indicates improvement in multi-label competency. Similar improvements in temporal consistency were observed for the remaining testing snippets which are shown in Figure \ref{fig:final_temp} alongside the outputs from video models.

\subsection{Multi-label Video Classifiers}
\begin{figure*}[htbp]
    \centering    \includegraphics[width=1.0\textwidth]{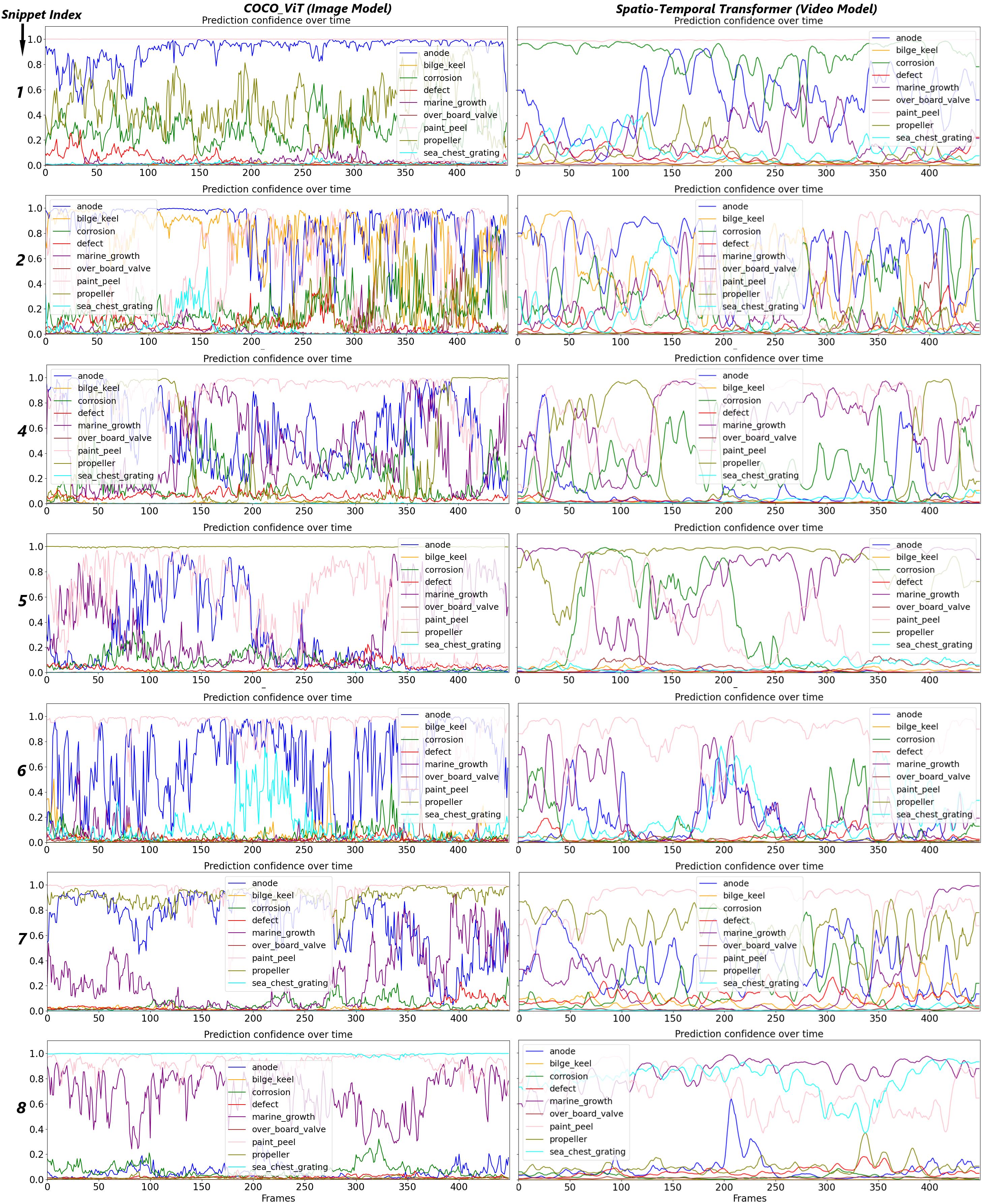}
    \caption{Temporal consistency comparison between the IMAGENET\_ViT and COCO\_ViT models on the snippets from Table \ref{tab:video_clips}.}
    \label{fig:final_temp}
\end{figure*}

Our initial attempt at implementing the video model utilizing uniform frame sampling did not result in convergence. Even after training for 1000 epochs, it exhibited a train and validation loss plateauing around 0.44. Also, the second variant using C3D features as tubelet embeddings did not yield a comparative performance. Nonetheless, our final approach produced promising results in terms of video classification performance. We trained 4 variants of video models within this approach by altering the weights of the spatial transformer encoder and the underlying feature pooling strategy for both the spatial and temporal transformers. Table \ref{tab:video_model_eval} outlines the performance evaluations of these video models on the validation video dataset. A detail of all the different training experiments is provided in the ablation study. Table \ref{tab:video_model_eval} indicates that model number 3 performs slightly better than the others. Accordingly, we have included the temporal observations of this model in Fig. \ref{fig:final_temp}, alongside the best image model. It is evident that the video model generates smoother temporal prediction confidence scores than the image model by stabilizing the predictions along the temporal dimension. While it has introduced some variance within the same class label, we discussed further improvement in the future work section which may overcome this limitation.

\begin{table}[htbp]
\caption{Evaluation metrics of video models on the validation dataset. ST = Spatial TRansformer, TT = Temporal Transformer, and Pool = Feature extraction.}
\begin{center}
\begin{tabularx}{\linewidth}{|c|c|X|X|X|X|X|X|X|}
\hline
 & \multirow{2}{*}{\textbf{Weights (ST)}} & \textbf{Pool (ST)} & \textbf{Pool (TT)} & \multirow{2}{*}{\textbf{Loss}} & \multirow{2}{*}{\textbf{Acc}} & \multirow{2}{*}{\textbf{Prec}} & \multirow{2}{*}{\textbf{Rec}} & \textbf{F1-score} \\
\hline
1 & COCO\_ViT & cls & cls & 0.30 & 0.59 & \textbf{0.78} & 0.72 & 0.69 \\
\hline
2 & IMAGENET\_ViT & cls & cls & 0.30 & 0.60 & 0.74 & 0.70 & 0.69 \\
\hline
3 & COCO\_ViT & cls & avg & 0.30 & \textbf{0.62} & \textbf{0.78} & \textbf{0.73} & \textbf{0.72} \\
\hline
4 & COCO\_ViT & avg & avg & \textbf{0.29} & 0.59 & \textbf{0.78} & 0.72 & 0.69\\
\hline
\end{tabularx}
\label{tab:video_model_eval}
\end{center}
\end{table}

\section{Ablation study}\label{analysis}
\subsection{Frame-based Video Classification}
To extract the best performance from image-based models for underwater ship hull inspection, several models were trained with gradual improvements by addressing the limitations of the LIACI dataset. The COCO 2014 dataset is a large-scale dataset that contains images with multiple object classes labeled in each image. In contrast, the IMAGENET dataset is primarily used for conventional image classification tasks where each image belongs to a single class. Hence, enabling our model to have multi-label classification capability, we initially train a ViT model on the COCO 2014 dataset using the hyperparameters and transformations mentioned in Table \ref{tab:init_param}.
\begin{table*}[htbp]
\caption{Analysis of different models \& results. FF = fully finetune \& PF = partial finetune.}
\begin{center}
\begin{tabularx}{\linewidth}{|c|c|c|c|X|X|X|X|X|X|X|X|X|X|}
\hline
&\multirow{2}{*}{\textbf{Model}} & \multirow{2}{*}{\textbf{Pretrain weight}} & \multirow{2}{*}{\textbf{\#Epochs}} & \multicolumn{2}{|c|}{\textbf{Loss}} & \multicolumn{2}{|c|}{\textbf{Accuracy}} & \multicolumn{2}{|c|}{\textbf{Precision}} & \multicolumn{2}{|c|}{\textbf{Recall}} & \multicolumn{2}{|c|}{\textbf{F1-score}} \\
\cline{5-14}
& & & & \textbf{Train} & \textbf{Val} & \textbf{Train} & \textbf{Val} & \textbf{Train} & \textbf{Val} & \textbf{Train} & \textbf{Val} & \textbf{Train} & \textbf{Val} \\
\hline
1& COCO ViT & IMAGENET 1K (FF) & 94 & 0.042 & 0.051 & 0.708 & 0.601 & 0.895 & 0.838 & 0.745 & 0.692 & 0.790 & 0.731 \\
\hline
2& LIACI ViT(initial) & IMAGENET 1K (FF) & 301 & 0.054 & \textbf{0.235} & 0.951 & 0.659 & 0.968 & \textbf{0.798} & 0.952 & 0.723 & 0.958 & 0.729 \\
\hline
3& LIACI ViT(initial) & COCO 2014 (FF) & 326 & 0.016 & 0.277 & 0.970 & 0.673 & 0.971 & 0.797 & 0.969 & 0.760 & 0.970 & 0.749 \\
\hline
4& LIACI ViT(extractor) & IMAGENET 1K (PF) & 277 & 0.267 & 0.281 & 0.593 & 0.565 & 0.764 & 0.741 & 0.648 & 0.621 & 0.672 & 0.642 \\
\hline
5& LIACI ViT(extractor) & COCO 2014 (PF) & 276 & 0.320 & 0.323 & 0.484 & 0.479 & 0.648 & 0.637 & 0.536 & 0.534 & 0.559 & 0.552 \\
\hline
6& LIACI ViT(step=5)& IMAGENET 1K (FF) & 99 & 0.263 & 0.288 & 0.597 & 0.556 & 0.778 & 0.740 & 0.651 & 0.606 & 0.678 & 0.632 \\
\hline
7& LIACI ViT(step=5) & COCO 2014 (FF) & 99 & 0.170 & 0.260 & 0.779 & 0.614 & 0.902 & 0.792 & 0.810 & 0.667 & 0.834 & 0.695 \\
\hline
8& LIACI ViT(step=50)& IMAGENET 1K (FF) & 99 & 0.018 & 0.277 & 0.971 & 0.672 & \textbf{0.972} & 0.808 & 0.971 & 0.739 & 0.971 & 0.744 \\
\hline
9& LIACI ViT(step=50) & COCO 2014 (FF) & 99 & \textbf{0.010} & 0.315 & \textbf{0.972} & 0.631 & \textbf{0.972} & 0.768 & \textbf{0.972} & 0.723 & \textbf{0.972} & 0.715 \\
\hline
10& LIACI ViT(final)& IMAGENET 1K (FF) & 99 & 0.034 & \textbf{0.235} & 0.961 & 0.674 & 0.969 & 0.805 & 0.962 & 0.753 & 0.964 & 0.747 \\
\hline
11& LIACI ViT(final) & COCO 2014 (FF) & 99 & 0.071 & 0.240 & 0.915 & \textbf{0.692} & 0.936 & 0.786 & 0.947 & \textbf{0.803} & 0.935 & \textbf{0.768} \\
\hline
\end{tabularx}
\label{tab:model_analysis}
\end{center}
\end{table*}
The COCO dataset consists of 82783 train and 40504 validation images and the model was trained for 94 epochs with a batch size of 16. We observed the model stops learning approximately after 30 epochs as both the training and validation losses become extremely low despite the accuracy still being confined under 0.7. Subsequently, we perform full finetuning of our two initial ViT models on the LIACI dataset. Table \ref{tab:model_analysis} includes the analysis of these initial models in rows 2 and 3, whereas row 1 corresponds to the COCO model. It is apparent from the F1-score or other metrics values of these two initial models that the ViT pre-trained on COCO performs better than the one pre-trained on IMAGENET. 

We investigated which model performs best in extracting features from the LIACI data. To devise this, we trained variants of the COCO and IMAGENET models using partial finetuning, where all the pre-trained weights except the classification part are frozen. The results are included in rows 4 and 5 of Table \ref{tab:model_analysis} which imply that the IMAGENET version outperforms the COCO model in feature extraction. However, the overall performance of the partial finetuning approach is still below the full finetuning approach. Therefore, we decided to keep the partial finetuning approach apart from our experiments. Additionally, we experiment with changing the optimizer from SGD to Adam with a weight decay of 0.3 to train both models but this led to a significant degradation in performance. We conducted experiments to explore the effects of different step sizes on the performance of the COCO and IMAGENET models. Along with the StepLR learning rate scheduler with a gamma value of 0.1 and test two more different step sizes: 5 and 50. To summarize, using a step size of 5 led to further regularization of the COCO model, but it also induced a decline in the overall performance for both models, as shown in rows 6 and 7 of Table \ref{tab:model_analysis}. On the other hand, the step size of 50 had a tendency to overfit the training for both models as assigned in rows 8 and 9. Finally, we deduced the best models with the configuration mentioned in the result section by considering both the quantitative evaluation measures and qualitative temporal performance which are also added in rows 10 and 11. COCO\_ViT is the best frame-based model which dominates all the validation evaluation metrics except the precision which is dominated by its counterpart IMAGENET\_ViT.

\subsection{Spatiotemporal-based Video Classification}
With the uniform frame sampling tokenization, we attempted to train our video models utilizing both image models and experimented with different learning rate schedulers. However, none of these approaches resulted in convergence during training. It is important to note that we were limited to using a dependent patch size to generate a total of 196 image patch embeddings from 7 frames, which were then fed into a ViT model. In addition to the video models discussed earlier, we also explored an approach that involved combining 3D CNN and ViT which we referred to as the tubelet embedding approach. Specifically, we extracted C3D features utilizing a pretrained 3D ResNet architecture and subsequently passed these features through our trained ViT-based image models. Although this approach resulted in convergence during training, the performance was not competitive enough to be included in the paper.

The spatial-temporal video model we reported in the paper has a total of 161.399M trainable parameters, with 75.600M of them belonging to the temporal transformer. Since we are utilizing a pre-trained ViT classifier as the spatial transformer, we freeze its weights during training and only update the weights of the temporal transformer, resulting in a substantial reduction in training time. One significant challenge that can contribute to poor performance is the limitation of transformers, which require pretraining on a large-scale dataset to optimize their performance. This is particularly relevant for the temporal transformer in our models, as its weights are initialized randomly, which can limit its ability to learn from the available data and lead to poor performance.


\section{Conclusion \& Future Work}\label{conclusion}
We have trained several multi-label ViT image classifiers and gradually improved them on the LIACI dataset to conduct framewise video inspections. In fact, the same trained model is also utilized during training multi-label video classifiers through different state-of-the-art approaches. However, while frame-based ViT classifiers are limited by their inability to capture temporal information, video classifiers can overcome this limitation by extracting both spatial and temporal features from the video. Spatial features are dominant in some videos, making image classifiers suitable for evaluation. Considering temporal features during video classification improves the robustness of the task, making it more effective for difficult video inspections like ours, and also stabilizes the model's prediction in the temporal dimension.

Although we conducted an exhaustive analysis, we believe there is still room for improving the performance of both image and video-based classifiers in an underwater environment. For example, exploring other pretraining strategies or designing custom architectures may yield better results. Additionally, gathering more diverse and high-quality data can also improve the performance of these models. Incorporating other techniques such as data augmentation, transfer learning, or ensembling can also be explored to improve the overall performance. Besides, introducing a quantitative metric to evaluate the temporal performance of the video-based classifiers would indeed be a useful research direction. By quantifying the temporal performance, we can have a more objective measure of how well the model is able to capture temporal information in the videos. This could potentially lead to further improvements in the model architecture or training process and ultimately result in better performance for video-based classification tasks in underwater environments.

Designing a new Vision Transformer architecture that is compatible with the uniform frame sampling tokenization of 7 frames could potentially overcome the convergence issue observed previously. Pretraining this new architecture on large-scale datasets before fine-tuning it for the LIACI dataset could also improve its performance. One significant challenge we faced is the limited size and weakly supervised nature of our video dataset. To address this, it is better to explore options such as acquiring a larger fully supervised dataset, using techniques like data augmentation and regularization to enhance generalization, or incorporating pretrained weights for the temporal transformer. By doing so, we could improve the robustness and effectiveness of our video inspection models.

In conclusion, we hope this work provides a benchmark for the development of image and video-based classifiers in underwater environments. The analysis will help researchers and developers to improve the accuracy and effectiveness of these classifiers and our findings will facilitate the application of these methods in real-world scenarios. Furthermore, we will also continue to focus on improving the video model and developing quantitative metrics to evaluate the temporal performance of video-based classifiers to improve their reliability and robustness.

\section*{Acknowledgment}
The authors express their gratitude to the collaborators within the LIACi project,
funded by the Research Council of Norway under project No 317854. The first author would like to thank the Erasmus Mundus MIR 
program funded by the European Union for providing his master's scholarship to study at the University of Toulon and the Norwegian University of Science and Technology. He also acknowledges Helene Schulerud, research manager of the Computer Vision group at SINTEF, for hosting him to conduct the master's thesis within the group.






\bibliographystyle{IEEEtran}
\bibliography{ref}

\end{document}